\documentclass[conference]{IEEEtran}
\usepackage{booktabs}
\usepackage{multirow}
\usepackage[table]{xcolor}
\usepackage{graphicx}
\usepackage{amsmath}
\usepackage{amssymb}
\usepackage{amsfonts}
\usepackage[most]{tcolorbox}
\usepackage{makecell}
\usepackage{array}
\usepackage{caption}
\usepackage{cite}

\ifCLASSINFOpdf
\else
\fi
\begin{document}
%
\title{Ghosts Beneath Textures: Texture-Relation Cues for Cross-Paradigm AI-Generated Image Detection}

\author{
\IEEEauthorblockN{
Haoyu Wang\IEEEauthorrefmark{4}\quad
Yiming Qin\IEEEauthorrefmark{4}\quad
Zhongjie Ba\IEEEauthorrefmark{4}\IEEEauthorrefmark{3}\quad
Ziping Dong\IEEEauthorrefmark{4}\quad
Jishen Zeng\IEEEauthorrefmark{2}\quad
Peng Cheng\IEEEauthorrefmark{4}\quad
Kui Ren\IEEEauthorrefmark{4}
}
\vspace{0.2em}
\IEEEauthorblockA{
\IEEEauthorrefmark{4}State Key Laboratory of Blockchain and Data Security, Zhejiang University\\
\IEEEauthorrefmark{2}Alibaba Group
\qquad
\IEEEauthorrefmark{3}Corresponding Author: zhongjieba@zju.edu.cn
}
}

\maketitle

\begin{abstract}
AI-generated images have proliferated rapidly, motivating extensive research. Most existing AI-generated image detectors are developed and evaluated under image-free generation paradigms, such as noise-based or text-guided generation.  However, image-conditioned generation has become increasingly important in practical applications, as it enables more fine-grained control over generated content. Detecting AI-generated images across these two paradigms creates a critical cross-paradigm detection problem that has long been overlooked. To study this problem, we construct \textbf{ConImageGen}, a benchmark for cross-paradigm AI-generated image detection. Evaluations on ConImageGen show that existing detectors fail to generalize reliably across image-free and image-conditioned generation.

To address this failure, this paper identifies a cross-paradigm forensic cue and provides a new perspective for generalized AI-generated image detection. Specifically, by suppressing semantic interference, we visualize, for the first time, semantics-irrelevant texture patterns across generation paradigms. These patterns exhibit structured local-global texture relations, indicating a generalizable form of forensic evidence. Motivated by this finding, we shift the focus from directly exploiting explicit artifacts to modeling texture relations and propose \textbf{DTS-Det}, a detection framework that captures and leverages such relations for generalized AI-generated image detection.

Extensive experiments validate the effectiveness of our method. DTS-Det achieves state-of-the-art performance across diverse evaluation settings, reaching \textbf{99.6\%} ACC on \textbf{ConImageGen} with a \textbf{10.5\%} gain over the best baseline. It also achieves \textbf{93.2\%}/\textbf{94.1\%} ACC in cross-dataset evaluation on PicoBanana/RAID and maintains detection rates of \textbf{95.2\%}/\textbf{88.1\%} under reconstruction attacks and black-box adversarial attacks, respectively.
\end{abstract}


%
\IEEEpeerreviewmaketitle

\section{Introduction}
\label{sec:introduction}
Media forensics plays an important role in social, legal, and economic security by providing technical support for assessing the authenticity of image content~\cite{verdoliva2020media}. With the rapid development of generative models, AI-generated images have become one of the major forms of image forgery, motivating extensive research on AI-generated image detection~\cite{aide,NPR,freqnet}. These studies aim to identify forensic evidence that distinguishes synthetic images from real ones.

\textbf{Forensic Cues under Image-Free Generation.}
Existing detectors mainly exploit two types of forensic cues: frequency-domain cues and spatial-domain cues~\cite{fenlei,zhang2019ganartifacts,dzanic2020fourier}. \textbf{Frequency-domain} methods characterize synthetic artifacts based on spectral responses. For example, F3Net~\cite{f3net} separates images into frequency-aware components and summarizes local frequency statistics, whereas FreqNet~\cite{freqnet} embeds learnable operations into the Fourier space to enhance high-frequency feature responses. These methods exploit abnormal spectrum-level regularities introduced during image synthesis. \textbf{Spatial-domain} methods instead analyze visual evidence in the image space, and can be further divided into semantics-related and semantics-irrelevant cues~\cite{fenlei}. \textbf{Semantics-related} methods leverage high-level content representations. For example, UnivFD~\cite{UnivFD} classifies images in the feature space of a pretrained CLIP~\cite{clip}
, and C2P-CLIP~\cite{c2pclip} introduces category-level common prompts into CLIP to enhance category-conditioned discrimination. AIDE~\cite{aide} further strengthens semantic representations by incorporating spatial artifact statistics from frequency-selected regions as complementary evidence. In contrast, \textbf{semantics-irrelevant} methods focus on low-level forensic traces beyond image content, such as the upsampling-induced pixel dependencies exploited by NPR~\cite{NPR} and the post-processing-related forensic traces modeled by TruFor~\cite{TruFor}.

However, these cue designs have been largely developed and evaluated under image-free generation paradigms, such as noise-based or text-guided generation, where images are generated primarily from random noise or textual prompts~\cite{ho2020ddpm,glide,rombach2022ldm}. In contrast, image-conditioned generation uses an input image as additional guidance, enabling fine-grained control over the generated content and making it increasingly important in practical applications~\cite{instructpix2pix,controlnet,dreambooth}. Nevertheless, whether existing detectors can generalize from image-free to image-conditioned generation remains unclear, revealing a critical cross-paradigm detection problem that has long been overlooked~\cite{ojha2023universal}.

\textbf{Cross-Paradigm Generalization Failure.}
A key prerequisite for studying cross-paradigm detection is a unified detection benchmark that covers both generation paradigms. Existing AI-generated benchmarks are mainly focus on image-free generation~\cite{GenImage,drct,raid}. Recent datasets such as X2Edit~\cite{x2edit} and PicoBanana~\cite{picobanana} include image-conditioned samples, but they are primarily intended for developing and evaluating image-conditioned generation models, making them unsuitable for direct use in cross-paradigm detection. To fill this gap, we introduce \textbf{ConImageGen}, a comprehensive benchmark dataset comprising 364K real and AI-generated images from 13 generative models, covering both image-free and image-conditioned generation paradigms. 

Based on this benchmark, we evaluate seven representative detectors from three families: frequency cues, semantics-related cues, and semantics-irrelevant cues. Our results show that image-conditioned samples are generally harder to detect than image-free samples, and cues learned under one generation paradigm often fail to generalize to the other. Even when trained with data from both generation paradigms, existing detectors improve only modestly. These results reveal the failure of existing detectors to handle cross-paradigm detection.

\textbf{Our work.}
In this paper, we analyze the failure modes of different cue families and identify semantics-irrelevant cues as a promising yet narrowly modeled source of cross-paradigm forensic evidence. We further visualize such texture patterns shared across models and paradigms, revealing new forensic evidence and offering a novel perspective for detection. This study is guided by the following research questions:

\textbf{RQ1:} What cues are shared across generation paradigms?

\textbf{RQ2:} How can such cues be identified and characterized as forensic evidence?

\textbf{RQ3:} How can such evidence be used to achieve generalizable detection?

\textbf{Semantics-irrelevant Texture Patterns Across Generation Paradigms.} To identify forensic cues shared across generation paradigms, we revisit the limitations of existing cue types. Frequency-domain cues are becoming increasingly difficult to exploit as generative models evolve~\cite{freq_fade,chandrasegaran2021closer}, while semantics-related cues may suffer from limited generalization when detectors rely on dataset-specific abnormal semantics or spurious correlations~\cite{Brea_Semantic,reduce_content}. Prior studies on compression-related, up-sampling-induced, and other low-level forensic clues suggest that semantics-irrelevant information can provide useful evidence for AI-generated image detection~\cite{NPR,TruFor}. However, these cues are often modeled around specific artifact sources or generation operations, limiting their role as shared evidence across generation paradigms. As a result, semantics-irrelevant information remains a promising yet narrowly modeled source of cross-paradigm forensic evidence. We therefore systematically examine whether semantics-irrelevant information itself contains forensic cues shared across generative models from different paradigms.

The first challenge is to reduce content interference, as image content can obscure underlying semantics-irrelevant texture patterns. Inspired by content suppression in Photo-Response Non-Uniformity (PRNU) analysis~\cite{PRNU}, we aggregate a large number of visually diverse samples to suppress image-specific components and make generation-related texture patterns more observable. Based on the visualization results, we show, for the first time, that semantics-irrelevant texture patterns exist in generative models across paradigms and manifest as diverse structural textures. These results suggest that such patterns may constitute a class of forensic cues shared across generation paradigms.

\textbf{From Texture Patterns to Texture Relations.} Compared with real images, generated images show more pronounced texture patterns in these visualizations, and their noticeable differences across generative models suggest that the observed patterns may be associated with generative model architectures. To validate this hypothesis, we adapt the PRNU~\cite{PRNU} extraction procedure to obtain single-image texture traces and aggregate them into model-level texture patterns. We then conduct peak-to-correlation energy (PCE)~\cite{PRNU} experiments to measure the correlations among these patterns. The results show that texture patterns are strongly associated with generative model architectures. However, as new generative models emerge, previously unseen texture patterns may undermine detectors that rely on explicit artifacts, leading to limited generalization~\cite{ojha2023universal,chandrasegaran2021closer}.

Motivated by these findings, we provide a new detection perspective that shifts the focus from directly exploiting explicit texture patterns to modeling texture relations, and introduce \textbf{semantics-irrelevant texture relations} as a new form of forensic evidence for cross-paradigm AI-generated image detection.

\textbf{A Comprehensive Framework for Generalizable Detection.} Based on the above analysis, we propose \textbf{DTS-Det}, a detection framework that models semantics-irrelevant texture relations for cross-paradigm AI-generated image detection. Specifically, we design a texture-relation encoder to capture structured dependencies between local and global textures, allowing the detector to exploit semantics-irrelevant cues beyond explicit texture patterns. We further introduce a relation-guided attention mechanism to guide the model toward more generalizable texture-relation representations. To enhance discriminative capacity, we incorporate a semantics-related stream that captures anomaly cues associated with high-level image content.

Extensive experiments validate the effectiveness and generalization ability of DTS-Det. On ConImageGen, DTS-Det achieves an accuracy of 99.6\%, outperforming the best-performing baseline by 10.5\%, and achieves state-of-the-art performance under multiple evaluation settings. In more challenging generalization scenarios, DTS-Det achieves the best performance under cross-dataset evaluation, with accuracies of 93.2\% on PicoBanana and 94.1\% on RAID. Under the cross-media setting, it achieves a video detection accuracy of 87.3\%, outperforming the best-performing baseline by 18.0\%. DTS-Det also remains robust to common image operations and adaptive evasion attacks, achieving detection rates of 95.2\% and 81.6\% under reconstruction-based and maximum-perturbation black-box adversarial attacks, respectively, outperforming the best-performing baseline by 29.5\% and 15.3\%.

\noindent\textbf{Contributions.}
This paper makes the following contributions:

\begin{itemize}
    \item We provide a novel and effective detection perspective that shifts from direct artifact exploitation to texture-relation modeling, motivated by our first observation that semantics-irrelevant texture patterns exist across generative models and paradigms.

    \item We construct \textbf{ConImageGen}, a benchmark covering two generation paradigms and 13 generative models. Based on this benchmark, we expose the generalization failure of existing detectors across generation paradigms.

    \item We propose \textbf{DTS-Det}, a detection network equipped with a texture-relation encoder and relation-guided attention, which explicitly models semantics-irrelevant texture relations as forensic evidence for cross-paradigm generalization.

    \item We conduct extensive experiments across nine generalization and robustness settings, covering challenging scenarios such as cross-media detection, reconstruction attacks, and black-box adversarial attacks. The results show that our method achieves state-of-the-art performance, demonstrating its effectiveness.
\end{itemize}

\section{Problem Formulation and Threat Model}
\label{sec:problem_threat_model}

\subsection{Problem Formulation}
\label{subsec:problem_formulation}

\textbf{Detection Task.} We formulate AI-generated image detection as a binary classification task. Given an input image $\mathbf{x} \in \mathcal{X}$, the goal is to determine whether it is real or AI-generated. We denote the label as $y \in \{0,1\}$, where $y=0$ and $y=1$ indicate real and AI-generated images, respectively. A detector $f_{\theta}: \mathcal{X} \rightarrow [0,1]$ predicts the probability that $\mathbf{x}$ is AI-generated, and the final prediction is obtained by applying a threshold $\tau \in [0,1]$:
\begin{equation}
    \hat{y} = \mathbb{I}\left[f_{\theta}(\mathbf{x}) \geq \tau \right].
\end{equation}

\subsection{Threat Model}
\label{subsec:threat_model}

\textbf{Deployment Assumption.}
We consider a practical deployment scenario where the detector receives only the image content as input. No provenance metadata, watermark information, source generator identity, or generation prompt is assumed to be available. The detector must therefore make decisions based solely on visual and forensic evidence contained in the image.

\textbf{Adversary Goal.}
The adversary aims to make an AI-generated image be classified as real while preserving its visual quality and semantic content. This corresponds to evasion attempts that weaken, overwrite, or perturb forensic traces without substantially changing the perceived image content~\cite{carlini2020evading,saberi2024robustness}.

\textbf{Adversary Capabilities, Attacks, and Scope.}
We consider a transfer-only black-box setting. In practical deployment, the adversary usually does not know which detector is used by the platform and has no access to the target detector's parameters, gradients, training data, or query outputs. However, the adversary may know common detector designs used in the community and can train or use external surrogate detectors to craft transferable attacks~\cite{papernot2017practical}.

Under this assumption, we evaluate two complementary evasion capabilities. First, the adversary may reconstruct the AI-generated image, using it as the input or reference in a reconstruction or image-conditioned re-generation pipeline~\cite{reconstructattack,drct}. This models practical image reuse, refinement, and re-dissemination scenarios, where the main semantic content is preserved but the original forensic traces may be rewritten or weakened.

Second, the adversary may conduct transferable black-box adversarial attacks~\cite{raid}. Since the target detector is inaccessible, the adversary crafts perturbations against external surrogate detectors and relies on transferability to attack the unseen target detector~\cite{carlini2020evading,papernot2017practical}. The target detector is excluded from the surrogate ensemble used for attack generation.

In our experimental evaluation, we additionally consider non-adaptive post-processing robustness under common image degradations to simulate routine image dissemination~\cite{wang2020cnn}. Our robustness claims are limited to the evaluated reconstruction-based attacks, transferable black-box adversarial attacks, and common degradation settings. We do not consider white-box attacks or fully adaptive attacks specifically optimized against the target detector, which require access to the deployed model or a different evaluation protocol and are beyond the scope of this work.


\section{Cross-Paradigm Analysis of Existing \\ Forensic Cues}
\label{sec:limitations_existing_cues}

In this section, we build a unified cross-paradigm analysis setting based on \textbf{ConImageGen} and complementary out-of-domain evaluations. We then revisit existing forensic cues to examine whether current detectors remain effective when transferring between image-free and image-conditioned generation paradigms. 

\subsection{Benchmark and Analysis Setting}
\label{secsub:benchmark}

\textbf{Construction of ConImageGen.} As discussed in Section~\ref{sec:introduction}, existing benchmarks leave a gap in cross-paradigm AI-generated image detection, as they do not provide a unified setting for evaluating detector generalization across image-free and image-conditioned generation paradigms. To fill this gap, we construct \textbf{ConImageGen}, a comprehensive benchmark covering both generation paradigms.

Specifically, the image-free subset is built upon GenImage~\cite{GenImage}, a million-scale AI-generated image detection benchmark that includes diverse image contents and multiple representative generators, including ADM\cite{adm}, BigGAN\cite{biggan}, Stable Diffusion v1.4\cite{stablediffusion}, Stable Diffusion v1.5\cite{stablediffusion}, VQDM\cite{vqdm}, GLIDE\cite{genvidbench}, Midjourney\cite{midjourney}, and Wukong\cite{wukonghuahua}. We derive the image-conditioned subset from X2Edit~\cite{x2edit} by selecting samples generated under real-image guidance, which aligns with the image-conditioned generation and detection setting. This subset covers five image-conditioned editing models, including Bagel\cite{bagel}, Kontext\cite{flux}, OmniConsistency\cite{omniconsistency}, GPT-4o\cite{gpt4o}, and Step1X-Edit\cite{step1x}. For each generator, we collect 10K training images, 2K validation images, and 2K testing images. In total, \textbf{ConImageGen} contains 182K AI-generated images and 182K corresponding real images, forming a binary classification benchmark with 364K images. This unified benchmark enables controlled evaluation of detector generalization within each paradigm and, more importantly, across image-free and image-conditioned generation.

\begin{figure*}[!htbp]
\centering
\includegraphics[width=\textwidth]{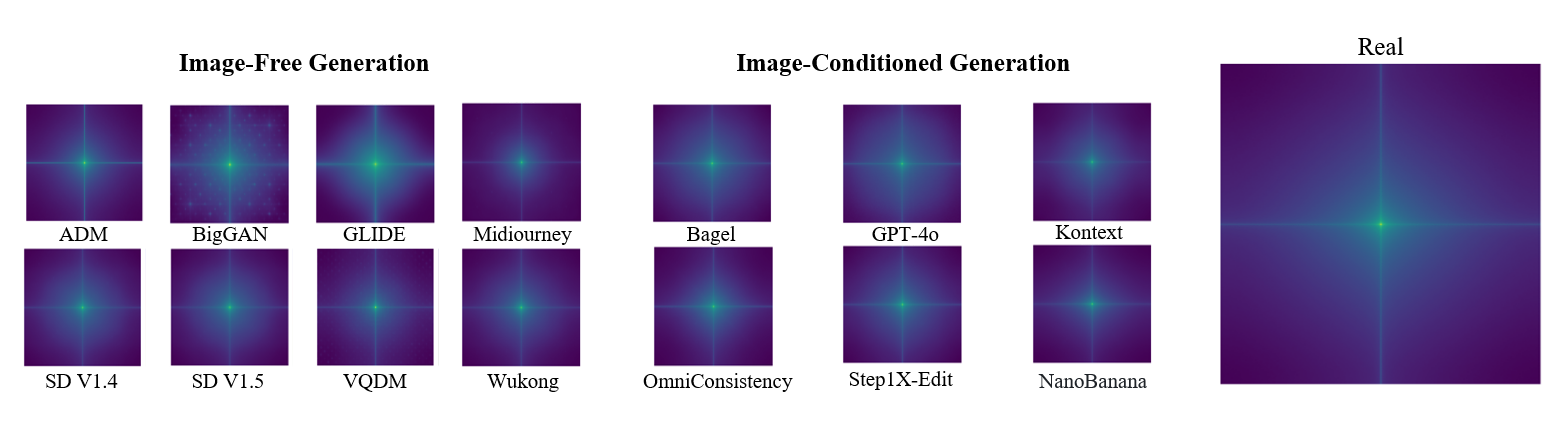}
\caption{Average spectra of real and generated images under different generation paradigms. Each spectrum is accumulated from 5,000 images. Some image-free models exhibit clear spectral cues, whereas the spectral differences between image-conditioned generated images and real images are substantially reduced.}
\label{fig:freq}
\end{figure*}

\textbf{Analysis Settings.}
Based on \textbf{ConImageGen}, we first define an in-domain setting. In this setting, detectors are \textbf{trained on samples from both image-free and image-conditioned generation} and tested on all generative models in the benchmark. This setting examines whether existing detectors can learn forensic cues that remain effective when both generation paradigms are covered during training. We report the average accuracy across all models as the overall metric.

We further evaluate out-of-domain generalization using image-free samples from RAID~\cite{raid} and image-conditioned samples from Pico-Banana~\cite{picobanana}. We use the same detectors \textbf{trained on both generation paradigms} in ConImageGen and directly test them on these external datasets to examine whether the learned forensic cues generalize beyond the training benchmark.

Finally, we define cross-paradigm settings to assess generalization across generation paradigms. Detectors are \textbf{trained on one paradigm} and tested on the other, covering both directions between image-free and image-conditioned generation. We report the average performance of the two directions as each method's cross-paradigm generalization ability.

Under these settings, we evaluate seven representative detectors from three cue families: frequency cues, semantics-related cues, and semantics-irrelevant cues.

\subsection{Revisiting Forensic Cues under Cross-Paradigm Evaluation}

In this subsection, we analyze the limitations of three types of forensic cues under cross-paradigm detection, based on our experimental results and findings from prior studies\cite{Brea_Semantic,freq_fade,reduce_content}. Specifically, frequency-domain cues are becoming increasingly difficult to exploit, semantics-related cues are sensitive to training data and generalize poorly when used alone, and semantics-irrelevant cues remain promising but narrowly modeled.

\begin{figure}[!htbp]
\centering
\includegraphics[width=0.45\textwidth]{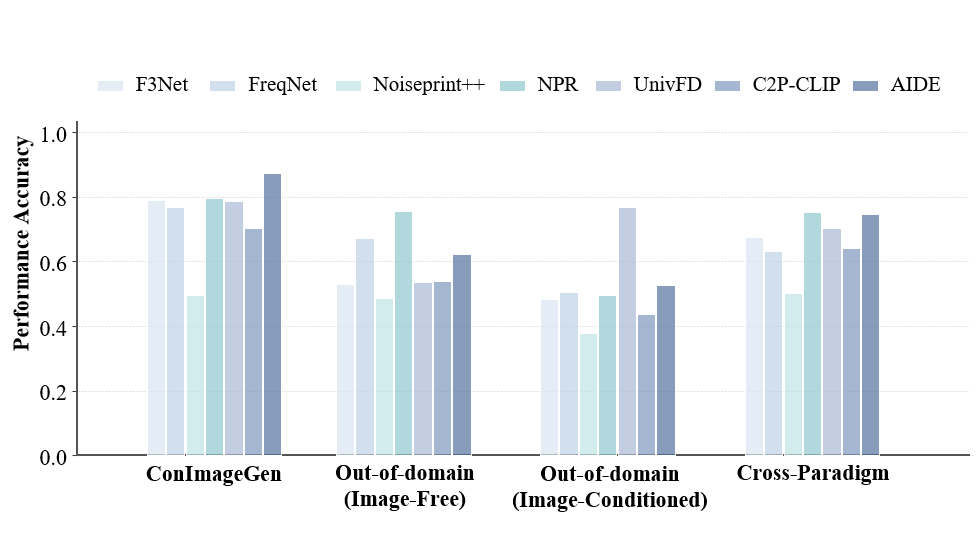}
\caption{Generalization performance of representative AI-generated image detectors. In-domain and out-of-domain accuracy are evaluated using detectors trained on both image-free and image-conditioned data; cross-paradigm accuracy is averaged over both transfer directions.}
\label{fig:cross}
\end{figure}

\textbf{Frequency cues.} For frequency-domain cues, we evaluate F3Net~\cite{f3net} and FreqNet~\cite{freqnet} as representative detectors. As shown in Table~\ref{fig:cross}, both methods perform moderately under the in-domain setting but degrade under cross-paradigm and out-of-domain evaluations, indicating that the learned spectral evidence is not stable across generation paradigms or data sources. This trend is consistent with prior studies showing that frequency artifacts are becoming increasingly difficult to exploit as generative models evolve~\cite{freq_fade}. Our accumulated spectrum visualizations in Fig.~\ref{fig:freq} further support this limitation: compared with image-free generation, image-conditioned generation exhibits weaker and less distinguishable spectral abnormalities. These results suggest that frequency-domain evidence is becoming less reliable for cross-paradigm detection.

\textbf{Semantics-related cues.} We evaluate UnivFD~\cite{UnivFD}, C2P-CLIP~\cite{c2pclip}, and AIDE~\cite{aide} as representative detectors for semantics-related cues. These methods achieve relatively strong in-domain performance, with AIDE performing especially well, but their performance varies substantially when the testing distribution changes. For example, their results fluctuate across cross-paradigm and out-of-domain settings, suggesting that such cues can be effective when training and testing data share similar content distributions but become less reliable under distribution shifts. This is consistent with prior studies showing that semantics-related detectors may exploit dataset-specific abnormal semantics or spurious correlations~\cite{Brea_Semantic,reduce_content}. Therefore, semantics-related cues provide useful discriminative information but are insufficient as standalone forensic evidence for cross-paradigm generalization.

\begin{figure*}[!t]
\centering
\includegraphics[width=\textwidth]{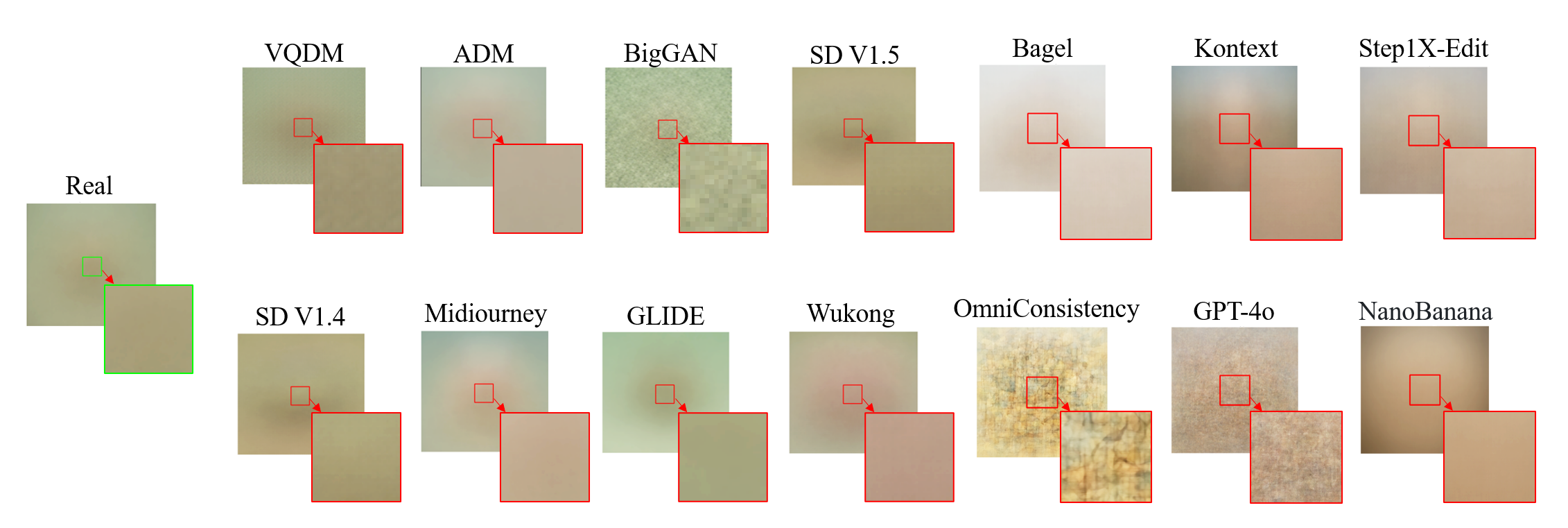}
\caption{Spatial visualization of semantically irrelevant texture patterns across 14 models from two generation paradigms. The results for each model are obtained from 5,000 images after suppressing the influence of semantic content. Different models exhibit distinctive texture patterns.}
\label{fig:pattern}
\end{figure*}

\textbf{Semantics-irrelevant cues.}
For semantics-irrelevant cues, we evaluate NPR~\cite{NPR} and Noiseprint++~\cite{TruFor} as representative methods that exploit low-level forensic clues beyond image semantics. NPR achieves competitive performance in some cross-paradigm settings, suggesting that semantics-irrelevant cues have potential for generalizable detection. However, its performance still degrades under out-of-domain evaluations, while Noiseprint++ remains unstable across most settings. These results suggest that although prior studies have explored compression-related, upsampling-induced, or other low-level forensic clues~\cite{NPR,TruFor}, such cues are often associated with specific artifact sources or post-processing operations. As a result, they may not fully capture the shared forensic evidence across different generative models and generation paradigms.

\label{subsec:cue_mismatch}

\begin{tcolorbox}[
    colback=gray!5,
    colframe=gray!45,
    boxrule=0.6pt,
    arc=2pt,
    left=4pt,
    right=4pt,
    top=4pt,
    bottom=4pt
]
\noindent\textbf{Finding.}
Existing forensic cues fail to generalize across paradigms, while semantics-irrelevant cues remain insufficiently modeled and require further exploration.
\end{tcolorbox}

\section{Texture Relation Across \\ Generation Paradigms}
\label{sec:texture_relation}

Building on the analysis in Section~\ref{sec:limitations_existing_cues}, we further explore semantics-irrelevant cues in this section. Specifically, we perform spatial visualization, extraction, and analysis of semantics-irrelevant texture patterns from generative models across different generation paradigms. The results show that semantics-irrelevant cues manifest as diverse, model-related texture patterns across generation paradigms. This analysis points to texture relation modeling as a novel detection perspective for cross-paradigm AI-generated image detection.

\subsection{Semantics-Irrelevant Texture Patterns}
\label{subsec:stable_traces}

The first step in studying semantics-irrelevant cues is to suppress content-related interference in images. In real-image forensics, stable content-independent traces can be estimated by extracting noise residuals and aggregating them from a large number of images via maximum-likelihood estimation~\cite{PRNU}. Inspired by this idea, we examine whether synthetic images also contain stable traces that are independent of image semantics. Given images of the same resolution generated by the same model, we aggregate a large number of samples from diverse categories to reveal model-level texture patterns while reducing image-specific content.

Formally, let $\mathcal{X}_g=\{\mathbf{X}^{(g)}_i\}_{i=1}^{N_g}$ denote a set of $N_g$ same-resolution images generated by model $g$, where 
$\mathbf{X}^{(g)}_i \in \mathbb{R}^{H \times W \times C}$. We compute the cross-sample average as follows:
\begin{equation}
\bar{\mathbf{X}}^{(g)}
=
\frac{1}{N_g}
\sum_{i=1}^{N_g}\mathbf{X}^{(g)}_i.
\label{eq:cross_sample_average}
\end{equation}

As discussed in prior work~\cite{PRNU}, when the samples span sufficiently diverse semantic categories, the semantic components at each spatial location tend to be weakly correlated across images and are therefore suppressed by averaging. In contrast, semantics-irrelevant texture traces that consistently appear across images generated by the same model can survive aggregation and become more visible in $\bar{\mathbf{X}}^{(g)}$. We refer to the resulting aggregated structure as a model-level texture pattern.  

Fig.~\ref{fig:pattern} shows that model-level texture patterns differ noticeably across generative models. Pronounced texture patterns can be observed in both image-free and image-conditioned generation, such as BigGAN in the image-free paradigm and OmniConsistency and GPT-4o in the image-conditioned paradigm. Stable Diffusion v1.4 and Stable Diffusion v1.5 exhibit similar texture patterns despite being different versions, suggesting that these patterns may be linked to the model architecture or generation mechanism. These patterns are not simple grid artifacts caused by resizing or JPEG compression, nor are they trivial repetitive textures~\cite{TruFor,NPR}. Instead, they exhibit complex model-related regularities, such as periodicity and symmetry. This observation may provide a possible explanation for the limited generalization of previous semantics-irrelevant cues. As these cues are often designed around specific operations or artifact sources, such as upsampling or compression~\cite{NPR,TruFor}, they may not fully capture structured model-level texture patterns and may consequently become less effective in these cases.

\begin{tcolorbox}[
    colback=gray!5,
    colframe=gray!45,
    boxrule=0.6pt,
    arc=2pt,
    left=4pt,
    right=4pt,
    top=4pt,
    bottom=4pt
]
\noindent\textbf{Finding.}
By suppressing semantic content, we visualize semantics-irrelevant texture patterns in the spatial domain. The visualizations reveal that generative models across different paradigms leave distinct semantics-irrelevant cues, which manifest as model-related texture patterns.
\end{tcolorbox}

\begin{figure}[!htbp]
\centering
\includegraphics[width=0.48\textwidth]{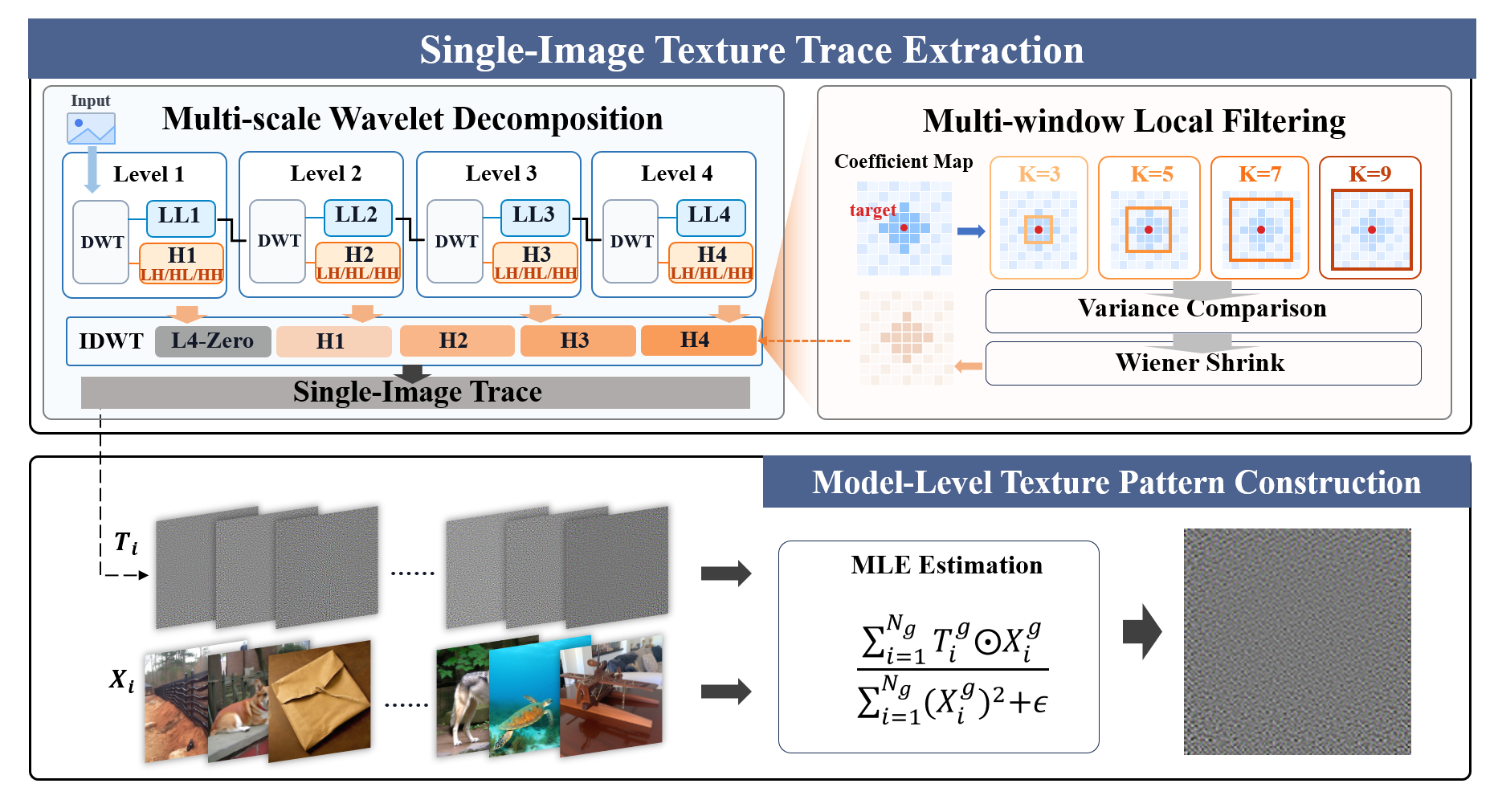}
\caption{Illustration of the proposed texture trace extraction pipeline. Each image is processed by multi-scale wavelet decomposition and multi-window local filtering to obtain a single-image trace, which is then aggregated via maximum-likelihood estimation to capture model-level texture patterns.}
\label{fig:trace}
\end{figure}

\subsection{Extraction of Texture Patterns}
\label{subsec:prnu_inspired_extraction}

This subsection extracts semantics-irrelevant texture patterns from AI-generated images and analyzes their characteristics. In AI-generated image detection, the detector typically receives a single image as input. We therefore first extract a single-image texture trace from each image to suppress semantic interference as much as possible. As illustrated in Fig.~\ref{fig:trace}, we then aggregate traces from images generated by the same model to construct a model-level texture pattern. This design allows image-specific content to be weakened while model-consistent texture structures are enhanced.

\textbf{Single-Image Texture Trace Extraction.}
Given an image $\mathbf{X}_i^{(g)}$ generated by model $g$, we follow wavelet-domain residual extraction with multi-window local variance estimation~\cite{PRNU,multiwindow}. We first perform a multi-scale wavelet decomposition using a $J$-level discrete wavelet transform:
\begin{equation}
\mathcal{W}\!\left(\mathbf{X}_i^{(g)}\right)
=
\left(
\mathbf{Y}_{L,i}^{(g)},
\left\{
\mathbf{Y}_{H,i}^{(g,j,b)}
\right\}_{j=1,\ldots,J;\,b=1,2,3}
\right),
\label{eq:dwt_decomp}
\end{equation}
where $\mathbf{Y}_{L,i}^{(g)}$ denotes the low-frequency approximation component, and $\mathbf{Y}_{H,i}^{(g,j,b)}$ denotes the high-frequency detail sub-band at scale $j$ and orientation $b$. Since image semantics are mainly concentrated in the low-frequency component, we discard $\mathbf{Y}_{L,i}^{(g)}$ and retain the high-frequency sub-bands as semantics-suppressed representations~\cite{PRNU}.

However, high-frequency components may still contain image-specific fluctuations~\cite{multiwindow}. To reduce unstable responses while preserving structured texture information, we estimate local statistics with multiple window sizes and apply a Wiener-style shrinkage operation to each high-frequency sub-band~\cite{PRNU,multiwindow}. Let $\Omega=\{3,5,7,9\}$ denote the set of local window sizes. For each $k\in\Omega$, we compute
\begin{equation}
\begin{aligned}
\boldsymbol{\mu}_{i,k}^{(g,j,b)}
&=
\mathcal{B}_{k}\!\left(\mathbf{Y}_{H,i}^{(g,j,b)}\right), \\
\boldsymbol{\nu}_{i,k}^{(g,j,b)}
&=
\mathcal{B}_{k}\!\left(\left(\mathbf{Y}_{H,i}^{(g,j,b)}\right)^2\right)
-
\left(\boldsymbol{\mu}_{i,k}^{(g,j,b)}\right)^2, \\
\boldsymbol{\nu}_{i,\min}^{(g,j,b)}
&=
\min_{k\in\Omega}
\boldsymbol{\nu}_{i,k}^{(g,j,b)} .
\end{aligned}
\label{eq:local_statistics}
\end{equation}
where $\mathcal{B}_{k}(\cdot)$ denotes a box filter with kernel size $k\times k$. The minimum variance across different window sizes provides a conservative estimate of local texture stability, which helps reduce content-dependent fluctuations while retaining structured responses that consistently appear in local neighborhoods.

Based on this estimate, we compute a shrinkage coefficient and obtain the filtered high-frequency sub-band:
\begin{equation}
\begin{aligned}
\mathbf{S}_{i}^{(g,j,b)}
&=
\frac{\boldsymbol{\nu}_{i,\min}^{(g,j,b)}}
{\boldsymbol{\nu}_{i,\min}^{(g,j,b)}+\sigma^2}, \\
\widetilde{\mathbf{Y}}_{H,i}^{(g,j,b)}
&=
\mathbf{S}_{i}^{(g,j,b)}
\odot
\mathbf{Y}_{H,i}^{(g,j,b)},
\end{aligned}
\label{eq:wiener_filtering}
\end{equation}
where $\sigma$ denotes the assumed noise standard deviation, and $\odot$ denotes element-wise multiplication. The filtered high-frequency components are then reconstructed into a semantics-suppressed residual, which is used as the single-image texture trace:
\begin{equation}
\mathbf{T}_{i}^{(g)}
=
\mathcal{W}^{-1}
\!\left(
\mathbf{0},
\left\{
\widetilde{\mathbf{Y}}_{H,i}^{(g,j,b)}
\right\}_{j=1,\ldots,J;\,b=1,2,3}
\right).
\label{eq:texture_trace}
\end{equation}
The resulting $\mathbf{T}_{i}^{(g)}$ suppresses low-frequency semantic content while retaining structured high-frequency responses, including periodic texture patterns that are consistent with the observed characteristics of synthetic images.

\textbf{Model-Level Texture Pattern Construction.}
Single-image texture traces still contain residual image-specific variations.Compared with raw images, the extracted traces contain a texture-pattern signal that remains weak and embedded in image-dependent noise~\cite{PRNU}; therefore, we prefer maximum likelihood estimation aggregation over simple accumulation to estimate a stable model-level pattern from trace responses extracted from images generated by the same model:
\begin{equation}
\hat{\mathbf{P}}^{(g)}
=
\frac{
\sum_{i=1}^{N_g}\mathbf{T}^{(g)}_{i}\odot \mathbf{X}^{(g)}_{i}
}{
\sum_{i=1}^{N_g}\left(\mathbf{X}^{(g)}_{i}\right)^2+\epsilon
},
\label{eq:model_pattern_mle}
\end{equation}
where $\epsilon$ is a small constant for numerical stability, and $\odot$ denotes element-wise multiplication. This aggregation reduces random noise that varies across individual images and enhances texture traces that consistently appear across samples from the same generator. The estimated $\hat{\mathbf{P}}^{(g)}$, shown in Fig.~\ref{fig:pattern}, is regarded as the model-level texture pattern of generator $g$ and is used for subsequent model-related analysis.

\subsection{Analysis of Model-Related Texture Patterns}
\label{subsec:model_specificity}

After extracting single-image texture traces and constructing model-level texture patterns, we further analyze whether these patterns are associated with generative model architectures. To this end, we use peak-to-correlation energy (PCE)~\cite{PRNU} to measure the match between a single-image texture trace and a reference model-level texture pattern.  The reference pattern is constructed from images generated by a candidate model and does not include the test image being evaluated. 

Given a test image $\mathbf{X}_i^{(g)}$ generated by model $g$ and a candidate model $h$, we define the query single-image texture trace and the reference model-level texture pattern as
\begin{equation}
\mathbf{T}_i^{(g)}
=
\mathcal{T}\!\left(\mathbf{X}_i^{(g)}\right),
\qquad
\mathbf{Q}^{(h)}
=
\hat{\mathbf{P}}^{(h)} ,
\label{eq:pce_trace_reference}
\end{equation}
where $\mathcal{T}(\cdot)$ denotes the single-image texture trace extraction operator, $\mathbf{T}_i^{(g)}$ is the query trace extracted from the test image, and $\mathbf{Q}^{(h)}$ is the reference model-level texture pattern of candidate model $h$. The test image $\mathbf{X}_i^{(g)}$ is not used when constructing $\mathbf{Q}^{(h)}$. We then compute the normalized correlation response between the query trace and the reference pattern:
\begin{equation}
\rho_i^{(g,h)}(u,v)
=
\frac{
\sum_{m,n}
\mathbf{T}_i^{(g)}(m,n)
\mathbf{Q}^{(h)}(m+u,n+v)
}{
\left\|\mathbf{T}_i^{(g)}\right\|_2
\left\|\mathbf{Q}^{(h)}\right\|_2
}.
\label{eq:pce_corr_map}
\end{equation}
A strong match between the query trace and the reference pattern should produce a sharp and well-localized peak in the correlation map. Let $(u_i^\star,v_i^\star)$ denote the peak location:
\begin{equation}
(u_i^\star,v_i^\star)
=
\arg\max_{u,v}
\rho_i^{(g,h)}(u,v).
\label{eq:pce_peak}
\end{equation}
We compute the average sidelobe energy by excluding a small neighborhood around the peak:
\begin{equation}
E^{(g,h)}_{i,\mathrm{sl}}
=
\frac{1}{|\Omega_{\mathrm{sl}}|}
\sum_{(u,v)\in\Omega_{\mathrm{sl}}}
\left[
\rho_i^{(g,h)}(u,v)
\right]^2,
\label{eq:pce_sidelobe}
\end{equation}
where $\Omega_{\mathrm{sl}}$ denotes the sidelobe region. The PCE score is then defined as the ratio between the squared peak response and the average sidelobe energy:
\begin{equation}
\operatorname{PCE}^{(g,h)}_i
=
\frac{
\left[
\rho_i^{(g,h)}(u_i^\star,v_i^\star)
\right]^2
}{
E^{(g,h)}_{i,\mathrm{sl}}
}.
\label{eq:pce_score}
\end{equation}
A higher PCE score indicates a stronger match between the query trace from model $g$ and the reference pattern of candidate model $h$~\cite{PRNU}. We treat PCE as a similarity score and evaluate its ability to distinguish same-generator pairs from different-generator pairs using AUC. Positive pairs correspond to $h=g$, while negative pairs correspond to $h\neq g$. A higher AUC indicates stronger association between query traces and the model-level texture patterns of their source generators.

The results show that texture traces usually match most strongly with the model-level texture pattern of their source generator. Stable Diffusion v1.4, Stable Diffusion v1.5~\cite{stablediffusion}, and Wukong\cite{wukonghuahua}, which are architecturally related according to prior work and open-source model analysis, also show stronger cross-associations than unrelated generators. This indicates that the extracted semantics-irrelevant texture traces contain model- and architecture-related information rather than random residuals.

This model-related property suggests that directly relying on specific semantics-irrelevant texture patterns may have limited generalization, as new generative models could introduce previously unseen patterns. Nevertheless, synthetic images from different generation paradigms still tend to differ from real images in the semantics-irrelevant texture space, indicating a potentially more generalizable forensic cue.

\begin{figure}[!ht]
\centering
\includegraphics[width=0.45\textwidth]{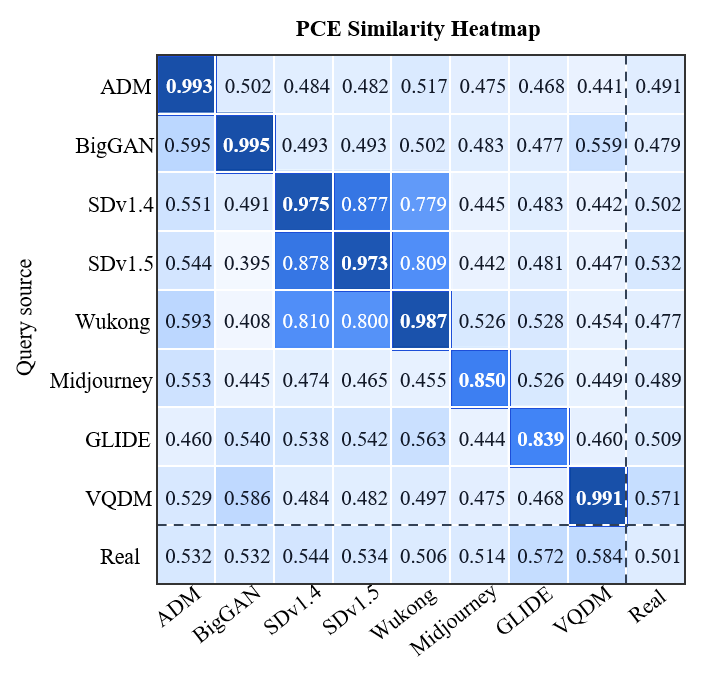}
\caption{PCE matching results between model-level texture patterns and image-level texture traces from different generative models. Strong diagonal responses and high PCE similarity among architecturally similar models reveal model-related characteristics in the extracted texture patterns.}
\label{fig:pce}
\end{figure}

\begin{tcolorbox}[
    colback=gray!5,
    colframe=gray!45,
    boxrule=0.6pt,
    arc=2pt,
    left=4pt,
    right=4pt,
    top=4pt,
    bottom=4pt
]
\noindent\textbf{Finding.}
By extracting and analyzing semantics-irrelevant texture patterns, we introduce texture relation modeling as a new perspective for AI-generated image detection.
\end{tcolorbox}

\begin{figure*}[!ht]
\centering
\includegraphics[width=\textwidth]{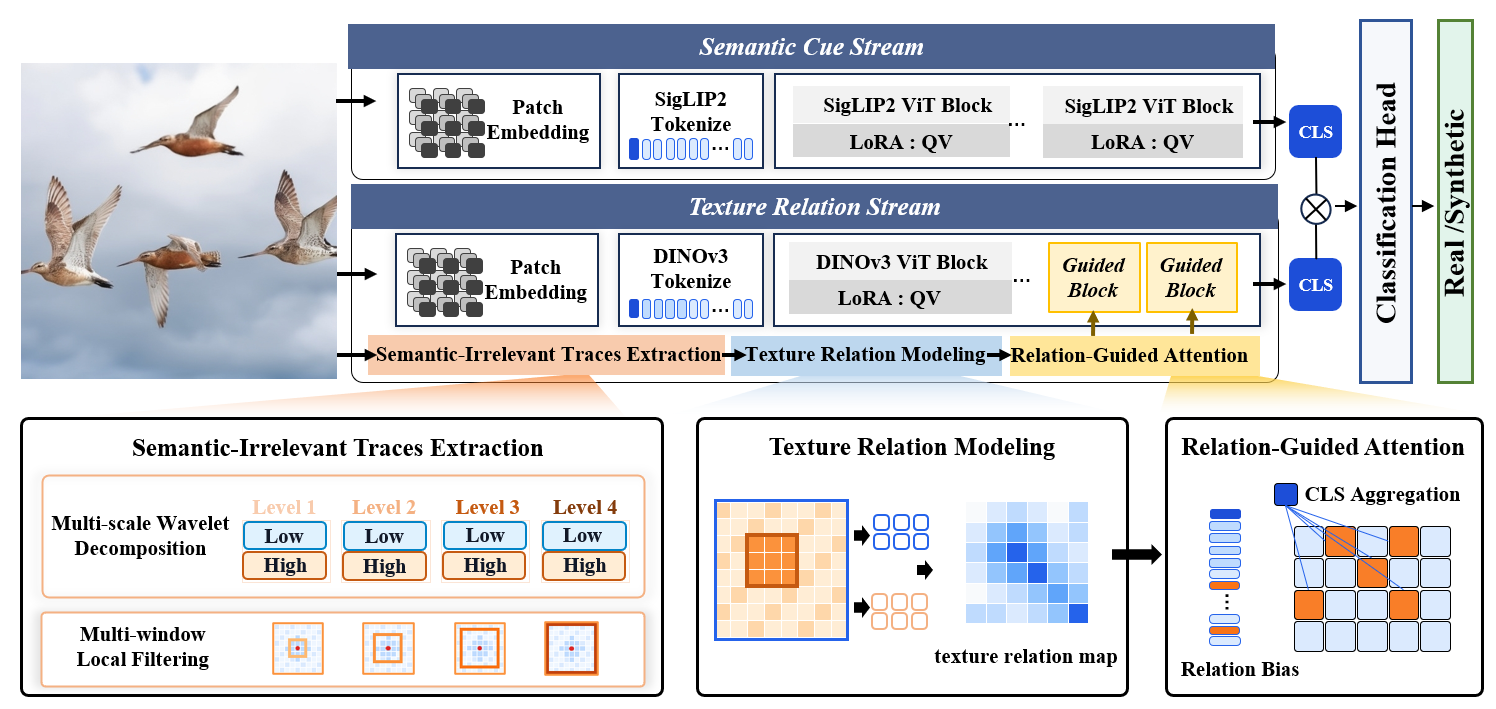}
\caption{Overview of the proposed DTS-Det framework. DTS-Det extracts semantics-irrelevant texture traces and models their relations through a texture-relation stream, where learned relation priors are injected into relation-guided attention blocks. A complementary semantic cue stream provides high-level semantic evidence for comprehensive AI-generated image detection.}
\label{fig:model_overview}
\end{figure*}

\section{The Proposed DTS-Det Framework}
\label{sec:method}

Motivated by the observations in Section~\ref{sec:limitations_existing_cues} and Section~\ref{sec:texture_relation}, we propose \textbf{DTS-Det}, a detection framework that leverages semantics-irrelevant texture relations and complements them with high-level semantics-related evidence for comprehensive AI-generated image detection.

\subsection{Overview}
\label{subsec:method_overview}

As illustrated in Fig.~\ref{fig:model_overview}, DTS-Det consists of three main components: a \textit{texture-relation stream}, a \textit{semantic cue stream}, and a \textit{prediction head}. The texture-relation stream takes semantics-irrelevant texture traces as input and provides a texture-relation prior that captures semantics-irrelevant forensic cues. The semantic cue stream operates on the original RGB image and extracts high-level semantics-related evidence. Finally, the prediction head aggregates the evidence from both streams and produces the detection result.

\subsection{Texture Relation Stream}
\label{subsec:texture_relation_stream}

Following the findings in Sec.~\ref{subsec:model_specificity}, the texture-relation stream shifts the focus from fixed texture traces to texture-relation modeling. It first suppresses semantic content and preserves semantics-irrelevant texture traces using the residual extraction operator introduced in Sec.~\ref{sec:texture_relation}. These traces are then encoded into a texture-relation prior, which characterizes relationships among texture patterns rather than fixed artifact appearances. The prior is injected into relation-guided attention to enable texture-relation learning for AI-generated image detection.

\subsubsection{Texture Relation Modeling}
\label{subsubsec:texture_relation_modeling}

The goal of texture relation modeling is to convert semantics-irrelevant texture traces into a token-level texture-relation bias for relation-guided attention. Given an input image $\mathbf{x}$, we first obtain its semantics-suppressed texture trace using the residual extraction operator $\mathcal{R}_{\mathrm{tex}}(\cdot)$. The resulting trace is then fed into a texture encoder $\mathcal{E}_{\mathrm{tex}}(\cdot)$ to capture texture organization beyond fixed artifact appearances.

We use SegFormer~\cite{segformer} as the texture encoder because its hierarchical structure can capture both local texture responses and broader texture organization. We use the encoder stage whose spatial resolution matches the patch-token layout of DINOv3~\cite{dinov3}, denoted as $\mathcal{E}_{\mathrm{tex}}^{a}(\cdot)$, and transform its output into a token-level texture-relation bias:
\begin{equation}
\mathbf{m}
=
\mathcal{P}_{b}
\left(
\mathcal{E}_{\mathrm{tex}}^{a}
\left(
\mathcal{R}_{\mathrm{tex}}(\mathbf{x})
\right)
\right),
\label{eq:relation_prior}
\end{equation}
where $\mathcal{P}_{b}(\cdot)$ denotes the bias transformation, which compresses the channel dimension of each patch feature into a scalar bias, applies normalization and bounded activation, and completes the bias sequence to match the full DINOv3 token layout. The resulting $\mathbf{m}$ provides patch-wise texture-relation evidence for subsequent relation-guided attention.

\begin{table*}[!ht]
\centering
\caption{In-domain performance comparison on ConImageGen. 
All methods are trained with data from both image-free and image-conditioned generation paradigms and evaluated on all generators. 
All values are reported as ACC/F1 (\%).}
\label{tab:indomain}
\setlength{\tabcolsep}{1.4pt}
\renewcommand{\arraystretch}{1.5}

\resizebox{\textwidth}{!}{
\begin{tabular}{lcccccccccccccc}
\toprule
\multirow{2}{*}{Baselines}
& \multicolumn{8}{c}{\textbf{Image-Free Generation}}
& \multicolumn{5}{c}{\textbf{Image-Conditioned Generation}}
& \multirow{2}{*}{AVG} \\
\cmidrule(lr){2-9}\cmidrule(lr){10-14}
& ADM & BigGAN & GLIDE & Midjourney & SDv1.4 & SDv1.5 & VQDM & Wukong
& Bagel & GPT-4o & Kontext & OmniCons. & Step1XEdit & \\
\midrule

ResNet50
& 68.8/69.6 & 54.1/54.6 & 68.6/69.9 & 78.0/77.7
& 79.2/79.6 & 82.7/83.0 & 62.8/62.0 & 77.7/78.3
& 56.0/57.0 & 64.3/63.5 & 53.1/51.4 & 72.5/72.5
& 64.6/63.0 & 67.9/67.8 \\

Xception
& 66.3/66.9 & 73.1/72.6 & 73.8/74.2 & 80.2/80.5
& 86.2/85.7 & 84.7/84.8 & 65.6/64.9 & 81.2/80.9
& 58.9/57.8 & 63.2/62.7 & 56.3/55.0 & 68.6/67.8
& 67.8/65.7 & 70.7/70.7 \\

\makecell[l]{SigLIP2}
& 61.0/55.2 & 54.2/42.0 & 73.5/72.8 & 85.4/86.9
& 85.6/88.9 & 86.6/88.7 & 70.1/68.1 & 85.4/87.2
& 50.1/59.8 & 57.7/69.1 & 49.2/65.1 & 65.4/74.0
& 55.2/64.9 & 67.6/70.8 \\

\makecell[l]{DINOv3}
& 81.5/77.6 & 97.7/97.8 & 92.2/91.7 & 86.8/85.2
& 97.9/97.9 & 97.9/97.9 & 96.9/96.8 & 98.4/98.4
& 60.7/47.5 & 78.6/80.3 & 52.9/41.1 & 88.0/89.1
& 65.5/64.3 & 84.2/82.0 \\

\makecell[l]{SigLIP2\\+DINOv3}
& 96.5/97.2 & 98.0/97.9 & 92.7/92.7 & 90.4/90.5
& 97.1/97.1 & 97.4/97.4 & 98.4/98.4 & 98.2/98.2
& 63.4/49.2 & 88.6/86.3 & 65.5/53.4 & 91.3/91.3
& 81.1/78.7 & 89.1/85.4 \\

\midrule
F3Net
& 83.7/83.3 & 85.9/86.3 & 83.6/84.0 & 90.2/90.1
& 92.9/93.1 & 90.7/90.9 & 70.0/69.8 & 91.4/91.2
& 55.0/52.8 & 78.2/78.4 & 52.1/43.7 & 87.5/86.6
& 64.0/60.6 & 78.9/78.2 \\

FreqNet
& 93.3/93.5 & 76.6/78.6 & 76.0/77.5 & 82.3/82.6
& 89.6/89.7 & 89.8/89.3 & 77.5/78.6 & 85.6/85.0
& 64.6/63.2 & 72.2/71.2 & 55.7/55.0 & 72.9/71.4
& 63.1/62.2 & 76.9/76.8 \\

\midrule
Noiseprint++
& 48.9/63.9 & 54.6/69.4 & 53.7/68.4 & 53.4/68.6
& 48.3/63.6 & 47.0/62.4 & 48.9/64.0 & 48.9/63.8
& 52.7/67.4 & 51.2/67.1 & 50.9/66.9 & 39.3/53.8
& 48.1/62.1 & 49.7/64.7 \\

NPR
& 93.3/93.5 & 78.8/79.5 & 81.4/82.0 & 84.2/83.5
& 81.7/82.3 & 83.3/83.2 & 83.5/83.3 & 83.9/84.1
& 59.6/57.8 & 88.4/88.0 & 64.3/62.1 & 87.5/87.6
& 65.6/65.4 & 79.7/79.4 \\

\midrule
UnivFD
& 77.4/71.8 & 94.6/94.5 & 94.2/94.1 & 76.6/71.7
& 88.2/87.1 & 90.2/89.7 & 87.7/86.3 & 83.9/81.7
& 56.7/39.6 & 70.3/71.2 & 56.0/46.2 & 88.1/89.9
& 58.9/51.1 & 78.7/74.9 \\

C2P-CLIP
& 71.6/65.0 & 91.1/91.4 & 90.4/90.8 & 74.8/71.0
& 74.4/68.9 & 76.0/72.8 & 76.7/75.4 & 74.4/69.4
& 53.7/61.9 & 57.7/65.0 & 52.4/56.4 & 65.2/73.1
& 56.7/66.8 & 70.4/71.4 \\

AIDE
& 97.6/97.6 & 85.4/86.4 & 76.0/76.4 & 89.9/89.8
& 97.7/97.7 & 97.6/97.6 & 94.0/93.9 & 97.8/97.9
& 82.7/80.2 & 77.4/77.2 & 50.4/43.5 & 97.2/97.3
& 91.1/90.9 & 87.3/86.6 \\

\midrule
\textbf{DTS-Det}
& \textbf{100.0/100.0} & \textbf{100.0/100.0} & \textbf{100.0/100.0} & \textbf{99.6/99.6}
& \textbf{100.0/100.0} & \textbf{100.0/100.0} & \textbf{100.0/100.0} & \textbf{100.0/100.0}
& \textbf{99.2/99.2} & \textbf{99.5/99.5} & \textbf{96.8/96.9} & \textbf{100.0/100.0}
& \textbf{99.5/99.5} & \textbf{99.6/99.5} \\

\bottomrule
\end{tabular}
}
\end{table*}


\subsubsection{Relation-Guided Attention}
\label{subsubsec:relation_guided_attention}

The token-level texture-relation bias $\mathbf{m}$ is injected into DINOv3 self-attention as a continuous attention bias~\cite{vaswani2017attention,liu2021swin}. Rather than using it as an independent classification feature, we use it to modulate DINOv3 token aggregation according to patch-wise texture-relation evidence. Let $\mathbf{X}^{(l)}$ denote the input tokens of the $l$-th DINOv3 block. Given the query, key, and value projections $\mathbf{Q}^{(l)}$, $\mathbf{K}^{(l)}$, and $\mathbf{V}^{(l)}$, the relation-guided attention output is computed as:
\begin{equation}
\hat{\mathbf{X}}^{(l)}
=
\operatorname{softmax}
\left(
\frac{
\mathbf{Q}^{(l)}
\left(
\mathbf{K}^{(l)}
\right)^{\top}
}
{\sqrt{d}}
+
\lambda
\mathcal{B}(\mathbf{m})
\right)
\mathbf{V}^{(l)},
\label{eq:relation_guided_attention}
\end{equation}
where $d$ is the head dimension, $\lambda$ is a learnable scaling factor, and $\mathcal{B}(\cdot)$ broadcasts $\mathbf{m}$ across attention heads and query tokens. In this way, the texture-relation bias is added along the key-token dimension, allowing patch tokens with stronger texture-relation responses to receive higher attention during token aggregation.

In practice, relation guidance is applied only to the last $L_g$ DINOv3 blocks, with $L_g=2$ by default. This introduces texture-relation evidence at the high-level token aggregation stage while avoiding unnecessary perturbation to lower-level pretrained representations.

\subsection{Semantic Cue Stream and Final Prediction}
\label{subsec:semantic_cue_stream}

Semantic-related features remain useful as complementary evidence for AI-generated image detection. Therefore, besides the texture-relation stream, DTS-Det introduces a semantic cue stream to provide complementary high-level evidence for comprehensive forensic analysis. The semantic cue stream is implemented with SigLIP2~\cite{siglip2}, a vision-language pretrained encoder. Given the original RGB image $\mathbf{x}$, this stream extracts a high-level semantics-related representation from image content. To support comprehensive forensic analysis, it complements the texture-relation stream with semantic evidence.

To adapt the pretrained encoders while preserving their generalizable representations, we use lightweight LoRA~\cite{lora} adaptation. Specifically, LoRA modules are inserted into the query and value projections of both pretrained encoders, enabling parameter-efficient adaptation without fully fine-tuning the backbones. For final prediction, we concatenate the class-token representations from the two streams. Let $\mathbf{c}_d$ denote the texture-relation-guided representation from the DINOv3 stream, and $\mathbf{c}_s$ denote the semantic representation from the SigLIP2 stream. The prediction is computed as:
\begin{equation}
\hat{y}
=
\sigma
\left(
W_c
\left[
\mathbf{c}_d ;
\mathbf{c}_s
\right]
+
b_c
\right),
\label{eq:final_prediction}
\end{equation}
where $[\mathbf{c}_d ; \mathbf{c}_s]$ denotes feature concatenation, $\sigma(\cdot)$ is the sigmoid function, and $W_c$ and $b_c$ are learnable classifier parameters. Through this fusion, DTS-Det combines semantics-irrelevant texture-relation evidence with semantics-related evidence, enabling a more comprehensive forensic analysis.

\section{Experiments}
\label{sec:experiments}

This section evaluates DTS-Det in terms of detection performance, generalization ability, and robustness under diverse evaluation scenarios. We further conduct ablation studies to analyze the contribution of each proposed component. Across these evaluations, DTS-Det consistently outperforms existing detectors. The results show that relation-guided texture modeling provides semantics-irrelevant evidence that generalizes across both paradigms, thereby improving the robustness and generalization of AI-generated image detection.

\subsection{Experimental Setup}
\label{subsec:experimental_setup}

\textbf{Benchmarks:}
We use ConImageGen, introduced in Sec.~\ref{secsub:benchmark}, for \textbf{in-domain} and \textbf{cross-paradigm} evaluation. It covers both image-free and image-conditioned generation with 13 generative models, using 2K generated images and 2K real images per generator for testing. For \textbf{cross-dataset} evaluation, we use PicoBanana~\cite{picobanana} and the clean split of RAID~\cite{raid}. From PicoBanana, we randomly sample 5K image-conditioned editing samples and their 5K corresponding source images as fake/real test pairs. From RAID, we randomly sample 1K clean generated images and 1K real images to evaluate generalization to unseen image-free generators. For \textbf{cross-media} evaluation, we use GenVidBench~\cite{genvidbench}, which covers 11 video generation models. We conduct frame-level evaluation by randomly sampling 1K generated videos and 1K real videos for each video generator, and extracting one random frame from each video as the image-level test sample. For \textbf{robustness} evaluation, we consider common image operations on ConImageGen, reconstruction-based attacks on ConImageGen, and transferable black-box adversarial attacks from RAID~\cite{raid}. For the RAID adversarial benchmark, we randomly sample 1K adversarial examples and their corresponding 1K real images for testing.

\begin{table}[!t]
\caption{Cross-paradigm generalization results. All values are reported as ACC/F1 (\%). IF and IC denote image-free and image-conditioned generation, respectively, and arrows indicate the training-to-testing direction.}
\label{tab:cross_paradigm}
\centering
\footnotesize
\setlength{\tabcolsep}{3pt}
\renewcommand{\arraystretch}{1.12}
\resizebox{\columnwidth}{!}{
\begin{tabular}{>{\centering\arraybackslash}m{1.30cm}p{2.35cm}ccc}
\toprule
Type & Baselines & \makecell[c]{IF$\rightarrow$IC} & \makecell[c]{IC$\rightarrow$IF} & AVG \\
\midrule

\multirow{5}{*}{\makecell[c]{Generic\\Classifier}}
& ResNet50 & 65.6/64.1 & 61.9/60.7 & 63.7/62.4 \\
& Xception & 70.7/70.1 & 61.9/62.7 & 66.3/66.4 \\
& SigLIP2-LoRA & 73.1/73.9 & 63.0/55.0 & 68.0/64.5 \\
& DINOv3-LoRA & 79.9/71.4 & 60.7/48.9 & 70.3/60.1 \\
& SigLIP2+DINOv3 & 84.5/83.8 & 84.5/81.0 & 84.5/82.4 \\

\midrule
\multirow{2}{*}{Frequency}
& F3Net & 72.3/72.8 & 63.0/63.5 & 67.6/68.1 \\
& FreqNet & 69.6/70.0 & 56.9/52.7 & 63.2/61.3 \\

\midrule
\multirow{2}{*}{\makecell[c]{Semantic-\\irrelevant}}
& Noiseprint++ & 50.4/69.6 & 49.9/66.3 & 50.2/67.9 \\
& NPR & 83.1/83.2 & 67.5/69.1 & 75.3/76.1 \\

\midrule
\multirow{3}{*}{Semantic}
& UnivFD & 77.2/79.7 & 63.5/62.8 & 70.4/71.3 \\
& C2P-CLIP & 68.8/74.4 & 59.4/53.2 & 64.1/63.8 \\
& AIDE & 81.7/81.6 & 67.9/66.2 & 74.8/73.9 \\

\midrule
& \textbf{Ours} & \textbf{92.2/90.8} & \textbf{96.2/95.9} & \textbf{94.2/93.3} \\
\bottomrule
\end{tabular}
}
\end{table}

\textbf{Baselines.}
We compare DTS-Det with representative detectors from different cue families: frequency-domain methods, including F3Net~\cite{f3net} and FreqNet~\cite{freqnet}; semantic-cue-based methods, including UnivFD~\cite{UnivFD}, C2P-CLIP~\cite{c2pclip}, and AIDE~\cite{aide}; and semantic-irrelevant methods, including Noiseprint++~\cite{TruFor} and NPR~\cite{NPR}. We further include generic classifier baselines, including ResNet50~\cite{resnet}, Xception~\cite{xception}, SigLIP2-LoRA~\cite{siglip2}, DINOv3-LoRA~\cite{dinov3}, and SigLIP2+DINOv3. ResNet50 and Xception are trained with a binary classifier, SigLIP2-LoRA and DINOv3-LoRA fine-tune the corresponding pretrained encoders with LoRA~\cite{lora} and a binary classifier, while SigLIP2+DINOv3 concatenates their class-token features for classification. These baselines help evaluate the effectiveness of texture-relation-guided modeling.

\textbf{Metrics and Implementation Details.}
We report accuracy and F1 score for all experiments, denoted as ACC/F1. 
Following common practice~\cite{NPR,aide}, all values are reported on a 0–100 scale in tables and result discussions. For DTS-Det, all input images are resized to $384 \times 384$, and the model is optimized using AdamW~\cite{adamw} with a learning rate of $1\times10^{-5}$ and a weight decay of 0.05. We use lightweight LoRA~\cite{lora} adaptation for the pretrained encoders. For a fair comparison, we follow the original preprocessing and training protocols of each baseline when available, and train and evaluate all methods using the same data splits.

\subsection{Main Results on ConImageGen}
\label{subsec:main_results_conimagegen}

We first evaluate all methods under the in-domain setting of \textbf{ConImageGen}, where detectors are trained with data from both image-free and image-conditioned generation paradigms and tested on all generative models in the benchmark. This setting examines whether a detector can learn forensic evidence that remains effective across both generation paradigms, rather than fitting cues that are specific to a particular paradigm.

As shown in Table~\ref{tab:indomain}, DTS-Det achieves the best overall performance, with an average ACC/F1 of \textbf{99.6/99.5} across 13 generative models. It improves over the best baseline results by 10.5/12.9 percentage points in ACC/F1. Since the generic backbone baselines already include SigLIP2, DINOv3, and their feature-level fusion, this gain cannot be explained by stronger pretrained semantic representations alone. Instead, it demonstrates the benefit of introducing semantic-irrelevant texture relations as complementary forensic evidence.

The performance breakdown further exposes the limitations of existing detectors. Although several baselines perform well on image-free generative models, they fail to maintain comparable performance on image-conditioned generative models, especially Bagel~\cite{bagel} and Kontext~\cite{flux}. This suggests that their learned evidence does not generalize well across different generation paradigms and may still depend on artifacts that are unstable or model-specific. In contrast, DTS-Det maintains consistently high performance across both generation paradigms. This further supports the effectiveness of modeling texture relations as more generalizable forensic evidence.

\begin{tcolorbox}[
    colback=gray!5,
    colframe=gray!45,
    boxrule=0.6pt,
    arc=2pt,
    left=4pt,
    right=4pt,
    top=4pt,
    bottom=4pt
]
\noindent\textbf{Result.}
DTS-Det maintains effective detection performance across both image-free and image-conditioned generation paradigms, suggesting that the learned texture relations provide useful forensic evidence for both paradigms.
\end{tcolorbox}

\begin{table}[!t]
\caption{Cross-dataset generalization results when training on ConImageGen. All values are reported as ACC/F1 (\%). PicoBanana and RAID are used as out-of-domain datasets for image-conditioned and image-free generation, respectively.}
\label{tab:cross_dataset}
\centering
\footnotesize
\setlength{\tabcolsep}{3pt}
\renewcommand{\arraystretch}{1.12}
\resizebox{\columnwidth}{!}{
\begin{tabular}{>{\centering\arraybackslash}m{1.30cm}p{2.35cm}ccc}
\toprule
Type & Baselines & PicoBanana & RAID & AVG \\
\midrule

\multirow{5}{*}{\makecell[c]{Generic\\Classifier}}
& ResNet50 & 57.3/55.3 & 46.1/38.8 & 51.7/47.1 \\
& Xception & 65.9/65.7 & 50.6/45.8 & 58.3/55.8 \\
& SigLIP2-LoRA & 59.2/63.5 & 45.4/48.2 & 52.3/55.9 \\
& DINOv3-LoRA & 52.0/25.1 & 56.9/54.5 & 54.5/39.8 \\
& SigLIP2+DINOv3 & 50.9/8.8 & 63.2/63.2 & 57.1/36.0 \\

\midrule
\multirow{2}{*}{Frequency}
& F3Net & 53.1/52.8 & 48.4/40.4 & 50.8/46.6 \\
& FreqNet & 67.1/66.5 & 50.6/45.2 & 58.9/55.9 \\

\midrule
\multirow{2}{*}{\makecell[c]{Semantic-\\irrelevant}}
& Noiseprint++ & 48.8/65.8 & 37.7/54.7 & 43.3/60.3 \\
& NPR & 75.6/75.9 & 49.6/45.4 & 62.6/60.7 \\

\midrule
\multirow{3}{*}{Semantic}
& UnivFD & 53.7/26.8 & 76.9/70.2 & 65.3/48.5 \\
& C2P-CLIP & 53.9/53.3 & 43.8/33.3 & 48.9/43.3 \\
& AIDE & 62.4/53.7 & 52.8/52.3 & 57.6/53.0 \\

\midrule
& \textbf{Ours} & \textbf{93.2/93.1} & \textbf{94.1/91.9} & \textbf{93.7/92.5} \\
\bottomrule
\end{tabular}
}
\end{table}

\begin{table*}[!ht]
\caption{Cross-media evaluation on GenVidBench. All detectors are trained on ConImageGen and evaluated at the frame level on videos generated by 11 different video generation models. All values are reported as ACC/F1 (\%).}
\label{tab:genvidbench}
\centering
\setlength{\tabcolsep}{1.4pt}
\renewcommand{\arraystretch}{1.5}

\resizebox{\textwidth}{!}{
\begin{tabular}{lcccccccccccc}
\toprule
\multirow{2}{*}{Baselines}
& \multicolumn{11}{c}{\textbf{Video Generation Models}}
& \multirow{2}{*}{AVG} \\
\cmidrule(lr){2-12}
& CogVideo & Keling & Mora & MuseV & Sora & SVD & ModelScope & OpenSora & Pika & T2V-Zero & VideoCraftV2 & \\
\midrule

ResNet50
& 53.5/52.3 & 58.5/58.7 & 61.5/61.7 & 54.0/53.5 & 50.5/49.2
& 56.5/55.8 & 58.0/58.4 & 62.5/63.7 & 57.0/56.7 & 61.5/62.8
& 61.5/61.7 & 57.7/57.6 \\

Xception
& 49.0/49.5 & 56.2/57.2 & 54.8/55.0 & 53.0/54.3 & 49.4/48.7
& 58.0/59.9 & 61.5/62.8 & 58.0/59.2 & 57.0/59.1 & 56.5/56.3
& 63.5/65.4 & 56.1/57.0 \\

SigLIP2-LoRA
& 50.5/61.7 & 48.2/59.0 & 44.5/55.0 & 47.5/58.4 & 40.3/49.7
& 48.3/59.6 & 56.5/65.0 & 50.5/58.0 & 60.5/69.2 & 50.7/58.2
& 56.0/66.7 & 50.3/60.1 \\

DINOv3-LoRA
& 76.5/80.9 & 61.5/75.5 & 57.5/81.0 & 57.5/58.3 & 54.6/54.8
& 56.0/57.3 & 77.0/81.3 & 75.5/79.8 & 75.0/79.3 & 76.2/80.3
& 76.8/81.3 & 67.6/73.6 \\

\makecell[l]{SigLIP2\\+DINOv3}
& 83.5/81.8 & 58.5/36.6 & 84.0/82.4 & 53.0/21.6 & 66.3/52.8
& 50.5/13.9 & 80.5/77.9 & 85.5/84.5 & 77.0/72.9 & 86.5/85.6
& 86.4/85.6 & 73.8/63.2 \\

\midrule
F3Net
& 57.5/57.7 & 52.0/51.9 & 58.5/57.8 & 49.0/47.4 & 48.4/47.6
& 53.5/52.7 & 56.0/56.4 & 56.5/56.7 & 54.0/54.9 & 56.2/55.6
& 53.8/53.5 & 54.1/53.8 \\

FreqNet
& 53.0/54.3 & 50.0/50.4 & 43.5/43.2 & 46.5/46.7 & 43.8/43.2
& 51.0/53.3 & 49.5/49.2 & 50.0/48.9 & 49.5/48.7 & 47.0/46.5
& 51.0/51.5 & 48.6/48.7 \\

\midrule
Noiseprint++
& 53.0/67.8 & 52.5/67.4 & 47.5/62.6 & 51.0/65.9 & 51.1/65.9
& 53.0/67.8 & 49.0/65.5 & 50.0/66.4 & 49.2/65.4 & 47.5/64.1
& 48.5/65.1 & 50.2/65.8 \\

NPR
& 49.5/50.2 & 58.0/59.9 & 58.5/61.0 & 52.5/54.9 & 54.5/55.2
& 52.0/55.6 & 66.5/68.2 & 61.5/63.2 & 55.0/57.5 & 56.2/56.9
& 63.8/66.3 & 57.1/58.9 \\

\midrule
UnivFD
& 81.0/83.0 & 66.5/65.6 & 75.5/76.9 & 52.0/42.1 & 69.4/69.0
& 53.0/44.0 & 69.0/70.4 & 77.5/80.2 & 73.0/75.2 & 76.5/79.1
& 69.0/70.4 & 69.3/68.7 \\

C2P-CLIP
& 49.0/59.8 & 46.0/56.4 & 48.5/59.3 & 49.5/60.4 & 49.5/60.2
& 47.5/58.2 & 52.0/63.1 & 40.5/49.8 & 53.0/64.1 & 38.0/46.5
& 47.2/57.6 & 47.3/57.7 \\

AIDE
& 55.8/65.1 & 52.2/60.2 & 61.5/67.3 & 53.0/61.7 & 47.9/55.6
& 58.2/67.5 & 58.5/73.7 & 58.4/63.7 & 52.0/59.6 & 62.5/68.1
& 54.0/58.8 & 55.8/63.6 \\

\midrule
\textbf{DTS-Det}
& \textbf{95.5/95.7} & \textbf{79.1/76.2} & \textbf{95.1/95.2} & \textbf{76.2/71.8} & \textbf{83.2/81.5}
& \textbf{76.1/71.7} & \textbf{92.2/92.6} & \textbf{91.5/92.1} & \textbf{88.5/89.0} & \textbf{92.1/92.5}
& \textbf{91.0/91.6} & \textbf{87.3/86.4} \\

\bottomrule
\end{tabular}
}
\end{table*}

\subsection{Generalization Evaluation}
\label{subsec:generalization_evaluation}

We evaluate the generalization ability of DTS-Det from three perspectives: cross-paradigm, cross-dataset, and cross-media. Cross-paradigm and cross-dataset settings examine whether the learned forensic cues are less dependent on specific generation paradigms, dataset sources, or generator families. Cross-media evaluation further provides a more challenging setting, where the detector is tested on a different media form, reflecting whether the learned cues are related to the visual-content synthesis process and can generalize beyond image-level training data. Across these settings, DTS-Det shows favorable generalization performance, validating the effectiveness of the proposed framework.

\subsubsection{Cross-Paradigm Generalization}
\label{subsubsec:cross_paradigm}

We first evaluate cross-paradigm generalization on ConImageGen by training detectors on one generation paradigm and testing them on the other. Specifically, IF$\rightarrow$IC trains on image-free generators and tests on image-conditioned generators, while IC$\rightarrow$IF trains on image-conditioned generators and tests on image-free generators. This setting directly tests whether the learned forensic evidence can transfer across generation paradigms, rather than relying on paradigm-specific artifacts.

As shown in Table~\ref{tab:cross_paradigm}, DTS-Det achieves the best performance in both directions, with an average ACC/F1 of 94.2/93.3. It outperforms the strongest baseline, SigLIP2+DINOv3, by 9.7/10.9 percentage points. The improvement is consistent in both transfer directions, reaching 92.2/90.8 for IF$\rightarrow$IC and 96.2/95.9 for IC$\rightarrow$IF. This shows that DTS-Det does not rely on the mixed training distribution in the in-domain setting, but learns cues that remain effective when the target paradigm is unseen.

The results also reflect the cue mismatch analyzed in Sec.~\ref{sec:limitations_existing_cues}. Frequency-based detectors degrade because spectral artifacts are not stable across paradigms. Semantic-based detectors provide partial transfer, but remain sensitive to the target paradigm. Existing semantic-irrelevant methods such as NPR are stronger than several semantic baselines, but still rely on narrowly modeled local traces. In contrast, DTS-Det models texture relations rather than exact trace templates, and combines them with semantic evidence. This directly supports our motivation that texture relations provide a more transferable semantic-irrelevant cue for cross-paradigm detection.

\subsubsection{Cross-Dataset Generalization}
\label{subsubsec:cross_dataset}

We next evaluate cross-dataset generalization by training all detectors on ConImageGen and testing them on out-of-domain data collected from PicoBanana\cite{picobanana} and the clean split of RAID\cite{raid}, without further fine-tuning. Specifically, we use PicoBanana to obtain out-of-domain image-conditioned samples, and use RAID-clean to obtain out-of-domain image-free samples. This setting tests whether the learned evidence can generalize to unseen data sources and generator families under both generation paradigms. As shown in Table~\ref{tab:cross_dataset}, DTS-Det achieves the best cross-dataset performance, with an average ACC/F1 of 93.7/92.5. On PicoBanana, DTS-Det reaches 93.2/93.1, outperforming the strongest baseline, NPR, by 17.6/17.2. On RAID-clean, DTS-Det obtains 94.1/91.9. 

The strong performance on both out-of-domain image-conditioned and image-free datasets shows that DTS-Det does not rely on ConImageGen-specific artifacts or generator-specific distributions. Instead, it indicates that texture relations provide generalizable semantic-irrelevant forensic evidence across datasets, generator families, and generation paradigms.

\begin{figure*}[!t]
\centering
\includegraphics[width=\textwidth]{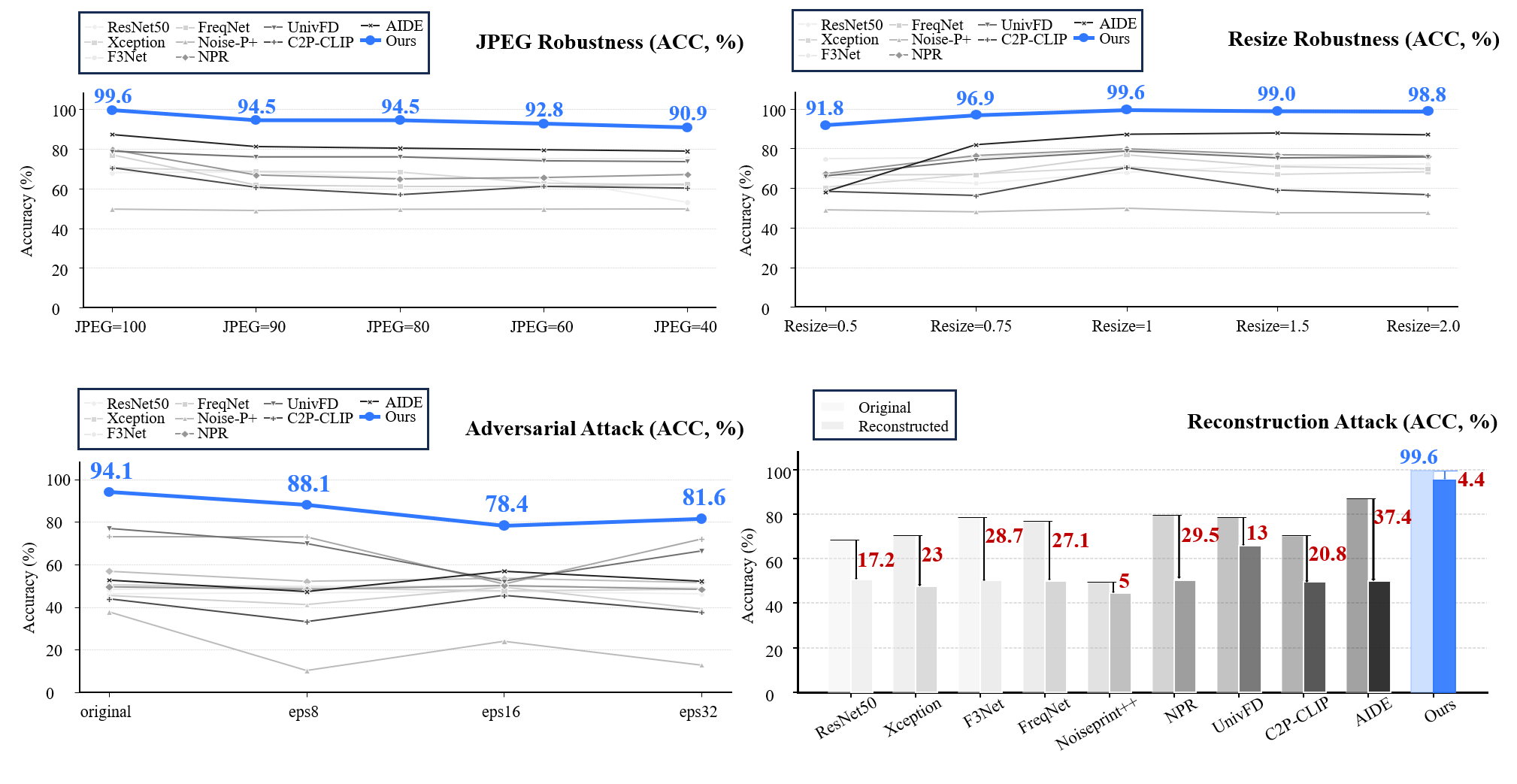}
\caption{Robustness comparison under image degradations and adaptive attacks. JPEG compression and resizing simulate real-world degradations, while adversarial and reconstruction attacks evaluate deployment robustness.}
\label{fig:robustness}
\end{figure*}

\subsubsection{Cross-Media Generalization}
\label{subsubsec:cross_media}

We further evaluate whether image-trained detectors can generalize to AI-generated video frames. All detectors are trained on ConImageGen and directly tested on frame-level samples from GenVidBench~\cite{genvidbench}. For each video generator, we randomly sample one frame from each selected generated video as a fake image, and apply the same strategy to real videos. No temporal information or video-level supervision is used, making this a challenging single-frame cross-media generalization setting. As shown in Table~\ref{tab:genvidbench}, DTS-Det achieves the best overall performance on generated video frames, with an average ACC/F1 of 87.3/86.4. It outperforms the strongest image-detector baseline, SigLIP2+DINOv3, by 13.5/23.2 percentage points, and achieves the best results across all tested video generators. 

Notably, DTS-Det is not trained on any video data, yet its single-frame ACC is even higher than the best Top-1 accuracy reported by many video-trained detectors on GenVidBench, although the evaluation protocols are not identical.

Since DTS-Det is trained only on images but remains effective on synthetic video frames, its learned evidence is not limited to still-image distributions. Instead, the result suggests that semantic-irrelevant texture relations are closely related to the synthesis behavior of generative models, making them a generalizable forensic perspective for detecting AI-generated visual content across media forms.

\begin{tcolorbox}[
    colback=gray!5,
    colframe=gray!45,
    boxrule=0.6pt,
    arc=2pt,
    left=4pt,
    right=4pt,
    top=4pt,
    bottom=4pt
]
\noindent\textbf{Result.}
The DTS-Det framework demonstrates strong generalization ability in both cross-paradigm and cross-dataset evaluations. Moreover, its zero-shot detection performance on synthetic videos further highlights the close connection between texture relations and the underlying synthesis behavior of visual-content generative models.
\end{tcolorbox}

\subsection{Robustness Evaluation}
\label{subsec:robustness}

We further evaluate whether DTS-Det remains reliable when forensic cues are weakened or adversarially perturbed. We consider two challenging scenarios: common image degradations and adaptive attacks.

\subsubsection{Common Image Degradations}
\label{subsubsec:image_degradation}

We evaluate the reliability of DTS-Det under common image operations on ConImageGen in the in-domain setting. Test images from all generation paradigms are perturbed with JPEG compression and resizing to simulate routine degradations during image sharing. All detectors are trained on clean images and tested under different JPEG quality factors and resizing scales. As shown in Fig.~\ref{fig:robustness}, DTS-Det achieves the best performance under both operations. Under strong JPEG compression with quality 40, it maintains 90.9 accuracy compared with 99.6 on clean images. Under the strongest downsampling scale of 0.5, it maintains 91.8 accuracy, while AIDE and C2P-CLIP drop to 58.1 and 58.2, respectively. These results show that DTS-Det remains reliable under common post-processing operations that degrade low-level artifacts, suggesting that local-global texture relations provide more stable semantics-irrelevant forensic evidence than fragile artifact-specific cues.

\subsubsection{Reconstruction-Based Attacks}
\label{subsubsec:reconstruction_attack}

We next evaluate robustness under reconstruction-based attacks. All detectors are trained on clean ConImageGen using data from all generators and then tested on reconstructed samples. Following the reconstruction protocol of DRCT~\cite{drct}, we construct this setting using AI-generated images from the Bagel subset of ConImageGen. This setting is challenging because reconstruction can rewrite the original forensic traces. As shown in Fig.~\ref{fig:robustness}, reconstruction severely weakens existing detectors. Frequency-based methods drop close to random performance, with F3Net decreasing from 78.9 to 50.2 and FreqNet from 76.9 to 49.8. Semantic-irrelevant methods are also fragile: NPR drops from 79.7 to 50.2. Even AIDE, the strongest clean baseline, decreases from 87.3 to 49.9. In contrast, DTS-Det remains stable, decreasing only from 99.6 to 95.2. This result directly supports our motivation: reconstruction can remove or rewrite exact artifact traces, but local-global texture relations provide more persistent semantic-irrelevant evidence when combined with semantic cues.

\subsubsection{Transferable Black-Box Adversarial Attacks}
\label{subsubsec:adversarial_attack}

Finally, we evaluate robustness against adaptive black-box adversarial attacks using RAID~\cite{raid}. Following the RAID protocol, detectors are tested on clean samples and adversarial samples under perturbation levels $\epsilon=8/255$, $16/255$, and $32/255$. Here, $\epsilon$ denotes the $L_{\infty}$ perturbation budget, which bounds the maximum per-channel pixel change after normalizing image intensities to $[0,1]$. Equivalently, these settings allow each pixel channel to be changed by at most 8, 16, or 32 intensity levels on the standard 8-bit image scale, with larger $\epsilon$ indicating stronger attacks. As shown in Fig.~\ref{fig:robustness}, adversarial perturbations severely degrade existing detectors. This effect is especially clear for semantic-dominant methods. SigLIP2+DINOv3 achieves 73.2 on clean RAID images but degrades substantially under adversarial perturbations. UnivFD similarly drops from 76.9 on clean images to 69.8 and 52.1 under $\epsilon=8/255$ and $\epsilon=16/255$, respectively. These results indicate that semantic forensic cues can be easily disrupted by adversarial perturbations.

DTS-Det shows stronger robustness under the same attacks, achieving 88.1 and 78.4 under $\epsilon=8/255$ and $\epsilon=16/255$, respectively. Even under the strongest perturbation level of $\epsilon=32/255$, it obtains the highest ACC of 81.6. These results indicate that adaptive adversarial attacks remain challenging, but texture-relation evidence reduces the model's reliance on vulnerable semantic cues.

\begin{tcolorbox}[
    colback=gray!5,
    colframe=gray!45,
    boxrule=0.6pt,
    arc=2pt,
    left=4pt,
    right=4pt,
    top=4pt,
    bottom=4pt
]
\noindent\textbf{Result.}
DTS-Det demonstrates robust performance across diverse evaluation scenarios, including common image degradations and adaptive attacks. Its resilience under various attack settings further indicates its practical potential for real-world AI-generated image detection.
\end{tcolorbox}

\subsection{Ablation Study}
\label{subsec:ablation}

\begin{figure}[!t]
\centering
\includegraphics[width=0.45\textwidth]{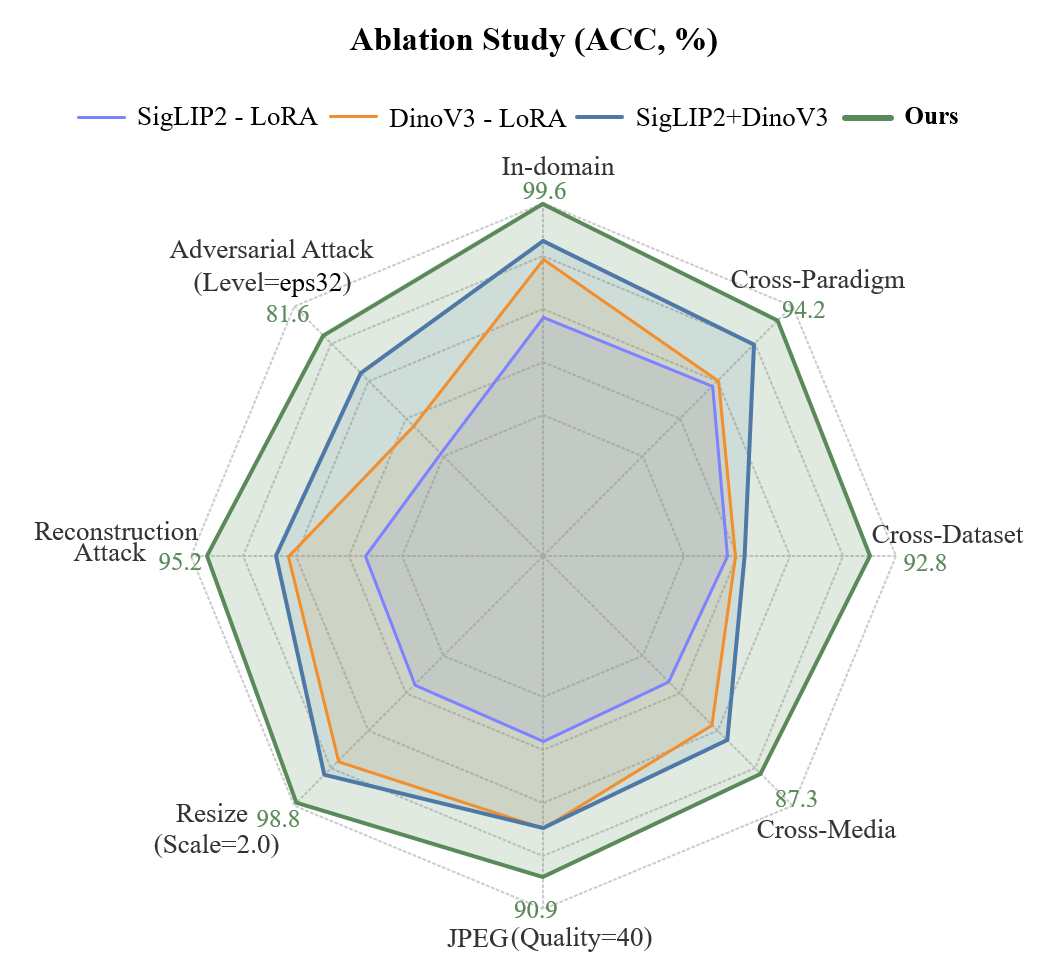}
\caption{Ablation comparison across in-domain, generalization, and robustness settings. 
Setting A uses SigLIP2, Setting B uses DINOv3, Setting C fuses the two semantic components, and Ours denotes the full DTS-Det with texture-relation guidance.}
\label{fig:ablation}
\end{figure}

We conduct ablation studies to verify the contribution of each component in DTS-Det. We compare four variants: (A) SigLIP2-LoRA, (B) DINOv3-LoRA, (C) feature-level fusion of SigLIP2 and DINOv3, and (D) the full DTS-Det with relation-guided texture modeling. All variants are evaluated under the same protocol across in-domain, generalization, and robustness settings. As shown in Fig.~\ref{fig:ablation}, single pretrained semantic encoders show unstable performance, while fusing SigLIP2 and DINOv3 improves the results, indicating their complementary semantic evidence. However, this semantic-only fusion still degrades under distribution shifts and attacks, showing that stronger semantic backbones alone are insufficient for robust forensic detection. With relation-guided texture modeling, DTS-Det achieves clear improvements across all settings, especially under cross-dataset transfer, reconstruction-based attacks, and transferable adversarial attacks. These results demonstrate the effectiveness of texture-relation guidance in improving detection performance.

\section{Conclusion}
\label{sec:conclusion}
Most existing AI-generated image detectors are developed under image-free generation, while image-conditioned generation is becoming increasingly important in practical applications. This paper investigated cross-paradigm AI-generated image detection, an important yet overlooked task that requires detectors to generalize between image-free and image-conditioned generation. 
We construct ConImageGen and show that existing detection cues fail to provide reliable cross-paradigm generalization. Through content suppression and texture analysis, we reveal semantics-irrelevant texture patterns shared across generation paradigms and further identify texture relations as forensic evidence for cross-paradigm generalization.
Based on this insight, we proposed DTS-Det, which models semantics-irrelevant texture relations and combines them with complementary semantic evidence. Extensive experiments across in-domain, cross-paradigm, cross-dataset, cross-media, and robustness settings demonstrate the effectiveness and generalization ability of DTS-Det. These findings highlight texture-relation modeling as a potential direction for cross-paradigm AI-generated image detection.

\section{Ethic Consideration}

This study aims to improve the reliability and generalization ability of AI-generated image detection, with a clear defensive purpose. Throughout the research, we follow principles of responsible data use and privacy protection. The study does not involve personal privacy, sensitive identity information, or security-sensitive data; all experiments are conducted on publicly available datasets and are used solely for training, evaluation, and analysis of detection methods.

\bibliographystyle{IEEEtran}
\bibliography{reference/reference}

\end{document}